\documentclass[journal]{IEEEtran}
\IEEEoverridecommandlockouts
\usepackage{cite}
\usepackage{amsmath,amssymb,amsfonts}
\usepackage{algorithmic}
\usepackage{algorithm}
\usepackage{graphicx}
\usepackage{textcomp}
\usepackage{xcolor}
\usepackage{colortbl}
\usepackage{tabularx}
\usepackage{seqsplit}

\usepackage{comment}
\usepackage{makecell} 
\usepackage{multirow}
\usepackage{flushend}
\usepackage{mathtools}
\usepackage{enumitem}
\usepackage[hyphens]{url}
\usepackage[breaklinks=true]{hyperref}
\usepackage{subfiles}
\usepackage{booktabs}
\usepackage{tikz} 
\usepackage{pifont}
\usepackage{cleveref}
\usepackage{balance}
\usepackage[dvipsnames]{xcolor}

\usepackage{todonotes}

\graphicspath{{Images/}}
\def\BibTeX{{\rm B\kern-.05em{\sc i\kern-.025em b}\kern-.08em
    T\kern-.1667em\lower.7ex\hbox{E}\kern-.125emX}}

\begin{document}

\title{Anomaly Detection on Resource-Constrained Devices with Reservoir Computing}

\author{Marco Angioli$^*$, Mathias Persson$^*$, Hiroto Tamura, Hadi Banaee, \\ Amy Loutfi, Gouhei Tanaka, Alberto Giaretta, Denis Kleyko
\thanks{$^*$Equal contribution. 
MA and DK acknowledge funding from the Swedish Strategic Research Foundation (Grant No. FFL24-0111).
HT acknowledges funding from the Japan Society for the Promotion of Science, JSPS KAKENHI (Grant No. JP24K20855), a project, JPNP14004, commissioned by the New Energy and Industrial Technology Development Organization (NEDO), and the Institute of AI and Beyond at the University of Tokyo.
AL and DK acknowledge funding from the Knut and Alice Wallenberg Foundation under the Wallenberg Scholars program (Grant No. KAW2023.0327).
GT acknowledges funding from the Japan Society for the Promotion of Science, JSPS KAKENHI (Grant No. JP25K24744). 
AG acknowledges funding from the Wallenberg AI, Autonomous Systems and Software Program (WASP) funded by the Knut and Alice Wallenberg Foundation.
AG and DK acknowledge funding from NordForsk (Grant No. 241658).
DK acknowledges funding from the Swedish Research Council (Grant No. 2025-05421).
}
\thanks{M. Angioli, M. Persson, H. Banaee, A. Loutfi, A. Giaretta, D. Kleyko are with the AI, Robotics and Cybersecurity Center (ARC) and the Department of Computer Science, Örebro University, 70281 Örebro, Sweden. A. Loutfi is also with the Department of Science and Technology, Linköping University, 58183 Linköping, Sweden; D. Kleyko is also with the Intelligent Systems Lab, RISE Research Institutes of Sweden, 16440 Kista, Sweden; (\mbox{e-mails}: \mbox{\{name.surname\}@oru.se}).}
\thanks{H. Tamura is with the Graduate School of Informatics, Osaka Metropolitan University, 558-8585 Osaka Japan and also with the International Research Center for Neurointelligence (IRCN), the University of Tokyo, 113-0033 Tokyo, Japan (\mbox{e-mail}: \mbox{h.tamura@omu.ac.jp}).    
}
\thanks{G. Tanaka is with the Graduate School of Engineering, Nagoya Institute of Technology, 466-8555 Nagoya, Japan and also with the International Research Center for Neurointelligence (IRCN), the University of Tokyo, 113-0033 Tokyo, Japan (\mbox{e-mail}: \mbox{gtanaka@nitech.ac.jp}).
}
}

\maketitle

\begin{abstract}
    Anomaly detection at the edge requires the deployment of models that achieve competitive performance, while minimizing computational cost and memory footprint.
    In this article, we investigate the suitability of the Mahalanobis Distance of Reservoir States (MDRS), an unsupervised time-series anomaly detection method, for resource-constrained devices and propose two optimizations to reduce its computational requirements. Specifically, we replace random reservoir connectivity (MDRS-Rand) with a circular shift (MDRS-Circ) and show that the anomaly score for an entire time series can be computed from the final reservoir state alone.
    Furthermore, we show that exploiting the reservoir's temporal memory enables more efficient input preprocessing, further reducing computational cost.
    We evaluate the MDRS variants on the MLPerf Tiny anomaly detection benchmark against the reference autoencoder and other anomaly detection methods, and deploy their inference kernels on a NUCLEO-F401RE microcontroller to assess latency, estimate energy consumption, and measure memory footprint. Among the evaluated methods, only MDRS and the autoencoder meet the benchmark's performance target, with MDRS requiring fewer operations and smaller models. With optimized preprocessing, MDRS achieves an ROC AUC of $0.869$, a model size of $24$ KB, an estimated energy consumption of $24$ µJ per inference, and a latency of $0.32$ ms, corresponding to a $75.9\times$ speedup over the autoencoder with standard preprocessing.
\end{abstract}


\section{Introduction}
\label{sec:intro}
IoT and resource-constrained devices are increasingly used to perform tasks at the edge. Many of such tasks fall into the category of anomaly detection applied to time-series~\cite{belay2023unsupervised}. For example, networking intrusion detection systems require methods that analyze incoming packets and infer whether they compose an ongoing cybersecurity attack. Another example is the analysis of environmental conditions (e.g., video or audio signals) to monitor processes and identify suspicious deviations from the normal conditions.
To achieve the near real-time performance necessary for carrying out these tasks successfully, computations need to be performed locally. Offloading them to the cloud is often not viable, as the communication overhead is non-negligible. However, given the limited computational capabilities of resource-constrained devices, it is crucial to develop computationally-efficient methods.

Tiny machine learning (tinyML)~\cite{tinyML_book} has emerged as an area specifically targeting these settings, enabling the deployment of ML models on resource-constrained devices with limited memory, storage, and computational capabilities.
To stimulate the development of methods for tinyML, the MLPerf Tiny benchmark~\cite{mlperf} provides a standardized framework for evaluating such models, including an anomaly detection task based on acoustic machine-condition monitoring. By defining target detection performance and measuring computational efficiency, the benchmark supports the evaluation of trade-offs between accuracy and deployment cost.

Reservoir computing is a family of randomized recurrent neural networks promising for tinyML~\cite{yan2024emerging}. In reservoir computing, most network weights are initialized randomly and remain fixed, and only the output weights are trained~\cite{RC09}. 
This architecture drastically reduces training cost compared to fully-trainable neural networks, making reservoir computing particularly suitable for resource-constrained deployment. Building on this foundation, Tamura et al. recently proposed the Mahalanobis Distance of Reservoir States (MDRS)~\cite{tamura2026distributional}, an unsupervised anomaly detection method that detects point-wise anomalies by measuring the deviation of reservoir states from an estimated distribution of normal conditions.  Unlike previous reconstruction- or prediction-based reservoir computing methods, MDRS produces temporally stable anomaly scores and avoids the need for supervised training of the output weights, making it a promising candidate for unsupervised anomaly detection on resource-constrained devices.

In this article, we investigate the suitability of MDRS for anomaly detection on resource-constrained IoT devices and propose two optimizations to reduce its computational requirements. First, we replace the random reservoir connectivity with a circular shift operation~\cite{RodanMinESN2011,Kleyko_2022}, eliminating the need for a dense recurrent matrix multiplication. Second, we introduce an end-only score that computes the Mahalanobis distance only for the final reservoir state rather than at every time step. We evaluate the MDRS variants on the MLPerf Tiny anomaly detection benchmark against the reference autoencoder and other anomaly detection methods (\Cref{sec:methods}) in terms of detection performance, computational cost, and memory footprint (\Cref{sec:exp:soft}). Among the evaluated methods, only MDRS and the reference autoencoder meet the MLPerf Tiny performance target, with MDRS requiring fewer operations and a smaller model. 
Moreover, since the reservoir integrates its input over time, MDRS can process single frames without the temporal stacking required by the autoencoder to meet the performance target, allowing further reductions in computational cost. When deployed on a NUCLEO-F401RE microcontroller (\Cref{sec:exp:hard}), MDRS achieves 6.9$\times$ lower latency and estimated energy consumption than the autoencoder with identical inputs. With optimized preprocessing, these improvements reach $75.9\times$ while meeting the MLPerf Tiny performance target with a model size of 24 KB. 

\section{Related work}
\label{sec:related}
A review of ML-based anomaly detection in IoT scenarios~\cite{IoT_Survey} emphasizes practical deployment constraints and the need to balance detection performance with computational cost, communication overhead, and privacy. It surveys approaches to IoT anomaly detection, comparing supervised, unsupervised, and semi-supervised learning methods. A complementary review focusing on unsupervised anomaly detection in multivariate IoT time-series has been presented in~\cite{belay2023unsupervised}. 

Statistical methods detect deviations from normal behavior using probabilistic models, distance measures, or outlier scores~\cite{moustafa2021dad}. Conventional ML methods, such as support vector machines (SVMs), decision trees, and clustering, learn decision boundaries or input transformations from data~\cite{liu2012isolation}. Deep learning methods use neural networks to model complex spatial or temporal dependencies~\cite{meidan2018n,shanmuganathan2023lstm}. 
Among deep learning methods, autoencoder-based methods have been deployed on resource-constrained devices within tinyML~\cite{tinyML_book}, including LSTM autoencoders for urban noise monitoring on ESP32 devices~\cite{hammad2023unsupervised} and compact autoencoders for machine fault detection on an Arduino Nano 33 BLE Sense~\cite{yap2024modified}. However, differences in datasets and hardware platforms make it difficult to compare their detection performance and computational cost. To enable such comparisons, the MLPerf Tiny benchmark~\cite{mlperf} provides a common anomaly detection task, a target performance, and a reference autoencoder. In its closed division, the submissions run this reference model on different hardware platforms and inference engines~\cite{mlperf_tiny_v13}. The open division also admits other models and preprocessing, but its few anomaly detection submissions used smaller or quantized fully-connected autoencoders on FPGAs~\cite{borras2022open} and on a RISC-V processor~\cite{mlperf_tiny_v07}.

In this study, we focus on reservoir computing~\cite{buonomano1995temporal, Jaeger2001, Maass2002}, a computationally efficient approach to time-series processing. 
Reservoir computing belongs to a class of randomized recurrent neural networks~\cite{scardapane2017randomness} in which input and recurrent weights are initialized randomly and fixed, so only the output weights are trained. This enables temporal modeling at a lower training cost than fully-trainable recurrent networks.
Existing methods for anomaly detection based on reservoir computing mainly rely on prediction or reconstruction errors~\cite{obst2008using,chang2009mote,chen2020imbalanced, lemos2021echo,devita2023esnae, kato2024reconstructive}. 
In contrast, MDRS (see \Cref{sec:methods:mdrs})~\cite{tamura2026distributional} estimates the distribution of reservoir states from normal data and uses the Mahalanobis distance from this distribution as the anomaly score. 
It, therefore, does not require explicit prediction or reconstruction of the input signal and is suitable for unsupervised learning settings. 
MDRS falls between conventional ML and statistical methods, and its properties make it a suitable candidate for anomaly detection in IoT scenarios. However, its detection performance, computational cost, and memory footprint have not been evaluated on widely used TinyML benchmarks nor demonstrated in resource-constrained deployments.

\section{Methods}
\label{sec:methods}

\subsection{MDRS: Mahalanobis Distance of Reservoir States}
\label{sec:methods:mdrs}
The MDRS method has two stages, a training stage and an inference stage~\cite{tamura2026distributional}. During training, normal data are passed through the reservoir of an Echo State Network (ESN)~\cite{Jaeger2001}, and the reservoir state distribution for the normal data is estimated. 
Assuming the reservoir states follow a multivariate Gaussian distribution, the parameters to be estimated are the mean vector and covariance matrix for the reservoir states.

During inference, the test data are fed into the ESN with the same parameters and transformed into a sequence of reservoir states. 
The Mahalanobis distance is then computed to evaluate how each reservoir state for the test data differs from the estimated reservoir state distribution for the normal data. 
The Mahalanobis distance serves as the anomaly score, where a higher score indicates a greater deviation from the estimated distribution of normal conditions and, therefore, suggests a higher likelihood of an anomaly. 
The reservoir-based input transformation enables nonlinear temporal feature extraction and dynamic state embedding, thereby enhancing the separability between normal and anomalous conditions in the reservoir state space~\cite{Maass2002}.

MDRS separates offline model estimation from online inference. Once the model has been trained, inference requires only the ESN configuration and the precision matrix of the normal reservoir-state distribution. Under the zero-mean assumption, no mean vector needs to be estimated. This small set of model parameters makes MDRS suitable for deployment on resource-constrained devices.

During training, the model estimates the structure present in the normal data. 
First, the training data are passed through the ESN, where the reservoir state is updated as follows:
\begin{equation}
    \label{eq:RunReservoir}
    \mathbf{x}(t+\Delta t) = \mathbf{x}(t) + \frac{\Delta t}{\tau} \left (-\mathbf{x}(t) + g\mathbf{J}\Phi (\mathbf{x}(t)) + h\mathbf{V}\mathbf{f}^{\mathrm{in}}(t) \right ),
\end{equation}
\noindent
where $\mathbf{x}(t)$ is the state of the $N$ reservoir neurons, $\mathbf{f}^{\mathrm{in}}(t)$ is the input, $\Phi$ is the component-wise hyperbolic tangent, and $\Delta t$, $\tau$, $g$, and $h$ are hyperparameters. The input projection $\mathbf{V}$ and the reservoir connectivity $\mathbf{J}$ matrices are random and fixed, and $\mathbf{x}(0) = \mathbf{0}$ at the start of each time-series.
The resulting reservoir states are then fitted to a multivariate Gaussian distribution. For simplicity, the normal reservoir-state distribution is assumed to be centered at the origin in this article. Accordingly, no mean vector is estimated, and the distribution is represented by its non-normalized covariance matrix (scatter matrix) $\mathbf{\Sigma}_0$, accumulated over the $K$ training time-series:
\begin{equation}
    \label{eq:sum}
    \mathbf{\Sigma}_0 := \sum_{k=1}^{K} \sum_{n=1}^{T_0 / \Delta t} \mathbf{x}_k(n\Delta t)\, \mathbf{x}_k(n\Delta t)^{\mathsf{T}},
\end{equation}
\noindent
where $\mathbf{x}_k(t)$ is the reservoir state for the $k$-th time-series and $T_0$ denotes the time-series duration.  
The matrix captures the spread and correlations of the normal reservoir states around the origin.
A regularization term is added before the matrix is inverted to obtain the regularized precision matrix $\mathbf{\Lambda}_0$:
\begin{equation}
    \label{eq:precision_matrix}
    \mathbf{\Lambda}_0 := (\mathbf{\Sigma}_0 + \alpha \mathbf{I}) ^{-1}.
\end{equation}
\noindent 
The regularization is performed as in ridge regression by including the positive parameter $\alpha$. 
During the inference stage, the resulting precision matrix characterizes the reservoir-state distribution, and a time-series is processed using the same reservoir dynamics as in the training stage, \Cref{eq:RunReservoir}. 
For a reservoir state at time $t = n \Delta t$, the anomaly score is computed using the squared Mahalanobis distance:
\begin{equation}
    \label{eq:MD}
    \delta_{\mathrm{M}} (\mathbf{x}(t), \mathbf{\Lambda}_0) := \mathbf{x}(t) ^{\mathsf{T}} \mathbf{\Lambda}_0 \mathbf{x}(t).
\end{equation}
Unlike the reconstruction error of an autoencoder, the Mahalanobis distance is a sum of contributions along the principal directions of the normal reservoir states, i.e., the eigenvectors of $\Sigma_0$, which shows along which directions a recording deviates from normal operation and could support explainable, human-in-the-loop monitoring.

\subsection{MDRS with simplified connectivity}
For the scope of this article, we refer to the standard MDRS with random reservoir connectivity, described in \Cref{sec:methods:mdrs}, as \texttt{MDRS-Rand}. 
In addition to \texttt{MDRS-Rand}, we investigate an MDRS with simplified reservoir connectivity based on the ideas from~\cite{RodanMinESN2011,Kleyko_2022}. 
Specifically, the randomly generated reservoir connectivity matrix $\mathbf{J}$ in \Cref{eq:RunReservoir} is replaced by a circular shift operation.
In this variant (denoted as \texttt{MDRS-Circ}), the reservoir neurons are connected in a ring, where each neuron is connected to a successor and the last neuron connects back to the first one. 
This connectivity can be represented by a permutation matrix containing a single non-zero value in each row and column. 

It is more efficient to replace the matrix-vector multiplication with the circular shift operation.
This eliminates the need to generate and store the random connectivity matrix $\mathbf{J}$. 
Consequently, this deterministic design reduces both the memory footprint and the computational cost of updating the reservoir, while allowing us to examine whether comparable anomaly detection performance can be maintained.

\subsection{Aggregation strategies for anomaly scores}
\label{sec:methods:aggregation}
In its original formulation, \texttt{MDRS-Rand} detects point-wise anomalies by computing the Mahalanobis distance for individual reservoir states (cf. \Cref{eq:MD}).
In contrast, the anomaly detection task in the MLPerf Tiny benchmark requires detecting anomalies at the time-series level~\cite{blazquez2021review}. 
Therefore, the sequence of point-wise anomaly scores must be aggregated into a single score representing the entire recording.

A similar aggregation problem arises in ESN-based time-series classification~\cite{prater2017comparison}. 
We consider two natural aggregation strategies. 
The first, called \emph{per-step aggregation}, computes the Mahalanobis distance for every reservoir state after the initial one, $\mathbf{x}(0) = \mathbf{0}$, and averages the resulting scores over all time steps. 
The second, called \emph{end-only score}, computes the Mahalanobis distance only for the reservoir state corresponding to the final time step of the recording. 
We also investigate intermediate variants of the per-step aggregation, in which the Mahalanobis distance is computed at a fixed stride $s$ rather than at every time step, i.e., at the time steps $s, 2s, \ldots , ns$.

\subsection{Baseline anomaly detection methods}
\label{sec:methods:baselines}
During the experimental evaluation in Sections~\ref{sec:exp:soft}-\ref{sec:exp:hard}, we compare the performance of the proposed MDRS variants with several baseline anomaly detection methods.   
First, the MLPerf Tiny benchmark includes a fully-connected autoencoder (\texttt{AutoEncoder}) as the reference model for the anomaly detection task~\cite{mlperf}. 
It includes full-precision weights in floating-point format (\texttt{AutoEncoder-FP32}; used for software evaluation) and weights quantized to $8$-bit integers (\texttt{AutoEncoder-INT8}; used for resource-constrained deployment). 
This model serves as the primary baseline.

Second, given our focus on resource-constrained deployment, we consider lightweight anomaly detection methods implemented in the widely used Python Outlier Detection (PyOD) library~\cite{zhao2019pyod}, with hyperparameters selected on the validation set: \texttt{GMM}, \texttt{PCA}, \texttt{HBOS}, \texttt{ECOD}, \texttt{COPOD}, \texttt{KNN}, \texttt{CBLOF}, \texttt{LODA}, \texttt{DeepSVDD}, \texttt{LSTMAD} (PyOD defaults, due to its training cost), and the time-series methods \texttt{SpectralResidual} and \texttt{MatrixProfile}. The last three methods process each recording as a sequence; the other methods score each input sample independently, and the anomaly score of a recording is the mean over its input samples.

\section{Data, preprocessing, and metrics}
\label{sec:data}

\subsection{Anomaly detection task in the MLPerf Tiny benchmark}
To evaluate the proposed MDRS variants, we use the anomaly detection task from MLPerf Tiny~\cite{mlperf}, an industry-standard benchmark suite for ML on resource-constrained devices. 
The task is based on the subset of the dataset from the DCASE2020 Challenge~\cite{koizumi2020description}, which comprises recordings of miniature machine operating sounds collected under both normal and anomalous conditions. 
Specifically, the MLPerf Tiny benchmark uses recordings from the \emph{toy-car} machine type. Each recording is stored as a WAV audio file, and the dataset is organized into separate training and testing sets.

The anomaly detection task is designed for unsupervised learning. During training, only recordings corresponding to normal machine operation are available, while the test set contains both normal and anomalous recordings. 
The training set contains 1\,000 recordings of normal operation for each of seven toy cars (7\,000 in total) mixed with environmental noise, whereas the test set contains unseen normal and anomalous recordings for four of these toy cars; following MLPerf Tiny, the test set consists of every tenth test recording of the DCASE2020 dataset (248 recordings).
Since the MLPerf Tiny benchmark does not define a validation set, we construct one using additional recordings from the same four toy cars, available in the DCASE2020 Challenge dataset but not included in the MLPerf Tiny test set. 
The validation set contains both normal and anomalous recordings and is used exclusively for hyperparameter selection. 
The MDRS model itself is trained only on normal data, while anomaly labels are used only to evaluate candidate hyperparameter settings.

The objective is to model normal operating behavior and assign an anomaly score to each recording, with larger scores indicating a higher likelihood of anomalies. This setting reflects realistic condition monitoring scenarios, where anomalies are rare and representative data are often unavailable.

\subsection{Preprocessing}
\label{sec:methods:preprocessing}
For reproducible comparisons across methods, the MLPerf Tiny benchmark adopts a standardized preprocessing. Each recording (sampled at 16\,kHz) is transformed into a log-mel spectrogram with $n_\text{mels} = 128$ mel bands, a window of 64\,ms (1024 samples), and a hop of 32\,ms (512 samples), of which the 200 central frames (frames 50-249) are kept. Five consecutive frames are concatenated into one vector, resulting in 196 input samples of 640 values per recording.  

We first adopt exactly this preprocessing to evaluate MDRS under the conditions of the benchmark (Sections~V-A and~V-B), and then we investigate how the preprocessing can be optimized for MDRS (Sections~V-C and~VI).

\subsection{Metrics}

MLPerf Tiny uses the area under the receiver operating characteristic curve (ROC AUC), hereafter referred to as AUC, to evaluate the anomaly detection performance. 
The receiver operating characteristic curve characterizes the trade-off between the true positive rate and false positive rate over all possible detection thresholds~\cite{ROC_reason}, while AUC summarizes this curve as a single scalar value.
Higher AUC values indicate better separation between normal and anomalous recordings.
We assign one anomaly score to each recording and compute the AUC over all recordings of the test set. Since the reservoir of MDRS is initialized randomly, we report its average AUC ($\pm$ standard deviation) over 50 random reservoir initializations.
For the anomaly detection task, MLPerf Tiny specifies the target performance of AUC$\geq 0.850$, which serves as the primary performance criterion in Sections~\ref{sec:exp:soft} and~\ref{sec:exp:hard}. 
From the perspective of the MLPerf Tiny benchmark, models achieving this target performance are considered equivalent in terms of anomaly detection performance, allowing computational efficiency to become the primary criterion for comparison.

In addition to anomaly detection performance, we evaluate the computational cost and memory footprint of the considered models. 
We quantify computational cost as the number of operations to process a recording during inference, providing a hardware-independent metric for comparison. For the software experiments (\Cref{sec:exp:soft}), we measure floating-point operations using the Performance Application Programming Interface (PAPI) library~\cite{terpstra2010collecting}, which accesses hardware performance counters. We use the \texttt{PAPI\_SP\_OPS} and \texttt{PAPI\_DP\_OPS} events to count single- and double-precision floating-point operations, respectively. Since these counters do not capture integer operations, \texttt{AutoEncoder-INT8}, which performs the same operations as \texttt{AutoEncoder-FP32} in 8-bit integer arithmetic, is assigned the same count as \texttt{AutoEncoder-FP32}. Only the inference is measured; the log-mel front-end, which is shared by all models, is excluded.
 
The memory footprint is characterized by model size and peak RAM usage during training and inference. Model size corresponds to the storage required by the saved model. Peak RAM usage is measured in a fresh Python process dedicated to each stage, with log-mel features precomputed. We report the increase in peak memory relative to memory usage before each stage, isolating the memory overhead attributable to the method from that of the Python runtime, libraries, and feature extraction front-end. Measurements are performed on an x86 host and reported as the median of five runs.

\section{Experimental evaluation in software}
\label{sec:exp:soft}

\subsection{Effect of reservoir size and connectivity on performance}
\label{sec:exp:size}

We begin by investigating how the reservoir size, $N \in \{32, 64, 96, 128, 192, 256\}$, and the reservoir connectivity (\texttt{MDRS-Rand} and \texttt{MDRS-Circ}) affect anomaly detection performance. 
For each reservoir size, a grid search is performed to determine the remaining hyperparameters ($\Delta t$, $\tau$, $g$, $h$, and $\alpha$). Table~\ref{tab:hyperparameters} lists the values
considered.

\begin{table}[t]
    \centering
    \renewcommand{\arraystretch}{1.2}
    \caption{Hyperparameter grid search values. Reservoir connectivity is evaluated for both \texttt{MDRS-Rand} and \texttt{MDRS-Circ}.}
    \label{tab:hyperparameters}
    \begin{tabular}{lp{6cm}}
        \toprule
        \textbf{Hyperparameter} & \textbf{Candidate values} \\
        \midrule
        $N$          & $\{32$, $64$, $96$, $128$, $192$, $256\}$ \\
        $\Delta t$   & $\{5{\times}10^{-5}$, $1{\times}10^{-4}$, $2{\times}10^{-4}$, $5{\times}10^{-4}$, $1{\times}10^{-3}\}$ \\
        $\tau$       & $\{0.001$, $0.005$, $0.01$, $0.02$, $0.03\}$ \\
        $g$          & $\{0.05$, $0.1$, $0.5$, $1$, $3\}$ \\
        $h$          & $\{0.05$, $0.1$, $0.5$, $1$, $3\}$ \\
        $\alpha$     & $\{0.001$, $0.005$, $0.01$, $0.02$, $0.03\}$ \\
        \bottomrule
    \end{tabular}
\end{table}

Throughout this experiment, we use the MLPerf Tiny preprocessing ($n_\text{mels}=128$, five stacked frames) and the per-step aggregation to obtain a single anomaly score for each recording. For each reservoir size, every configuration is first evaluated with \texttt{MDRS-Rand} and a single random initialization, and the ten configurations with the highest AUC on the validation set are evaluated further over $50$ initializations; the configuration with the highest average validation AUC is selected and used for both \texttt{MDRS-Rand} and \texttt{MDRS-Circ}. For $N = 64$, the reservoir size used in the following, this yields $\Delta t = 5\times10^{-5}$, $\tau = 0.005$, $g = 3$, $h = 0.1$, and $\alpha = 0.001$. 

\Cref{tab:auc_vs_N} reports the average AUC on the test set for different reservoir sizes. Both variants achieve the MLPerf Tiny target performance from $N = 64$ upwards, whereas the smallest reservoir ($N = 32$) falls just below it ($0.843$ and $0.836$). The highest AUC is obtained with $N = 128$ ($0.894$ and $0.896$), and larger reservoirs do not improve it further, possibly because the amount of training data remains fixed while the number of parameters of the precision matrix grows quadratically with~$N$. We use $N = 64$ in the following, as it is within $0.02$ of the highest AUC at about half the cost of $N = 128$ (\Cref{tab:flops}).  

\begin{table}[t]
\centering
\footnotesize
\setlength{\tabcolsep}{8pt}
\renewcommand{\arraystretch}{1.3}
\caption{Average AUC ($\pm$ std. deviation over 50 random initializations) on the test set as a function of reservoir size, $N$, with the per-step aggregation and the MLPerf Tiny preprocessing.}
\label{tab:auc_vs_N}
\begin{tabular}{c|rr}
  \textsc{Reservoir}, $N$ & \texttt{MDRS-Rand} & \texttt{MDRS-Circ} \\
  \midrule
  $32$ & 0.843 $\pm$ 0.021 & 0.836 $\pm$ 0.021 \\
  \rowcolor{gray!12}
  $64$ & 0.878 $\pm$ 0.012 & 0.873 $\pm$ 0.012 \\
  $96$ & 0.882 $\pm$ 0.009 & 0.882 $\pm$ 0.010 \\
  \rowcolor{gray!12}
  $128$ & 0.894 $\pm$ 0.004 & 0.896 $\pm$ 0.005 \\
  $192$ & 0.890 $\pm$ 0.004 & 0.890 $\pm$ 0.004 \\
  \rowcolor{gray!12}
  $256$ & 0.883 $\pm$ 0.004 & 0.884 $\pm$ 0.005 \\
  \bottomrule
\end{tabular}
\end{table}

\subsection{Comparison of aggregation strategies}
\label{sec:exp:aggregation}
\begin{figure}[t]
    \centering
    \includegraphics[width=1\linewidth]{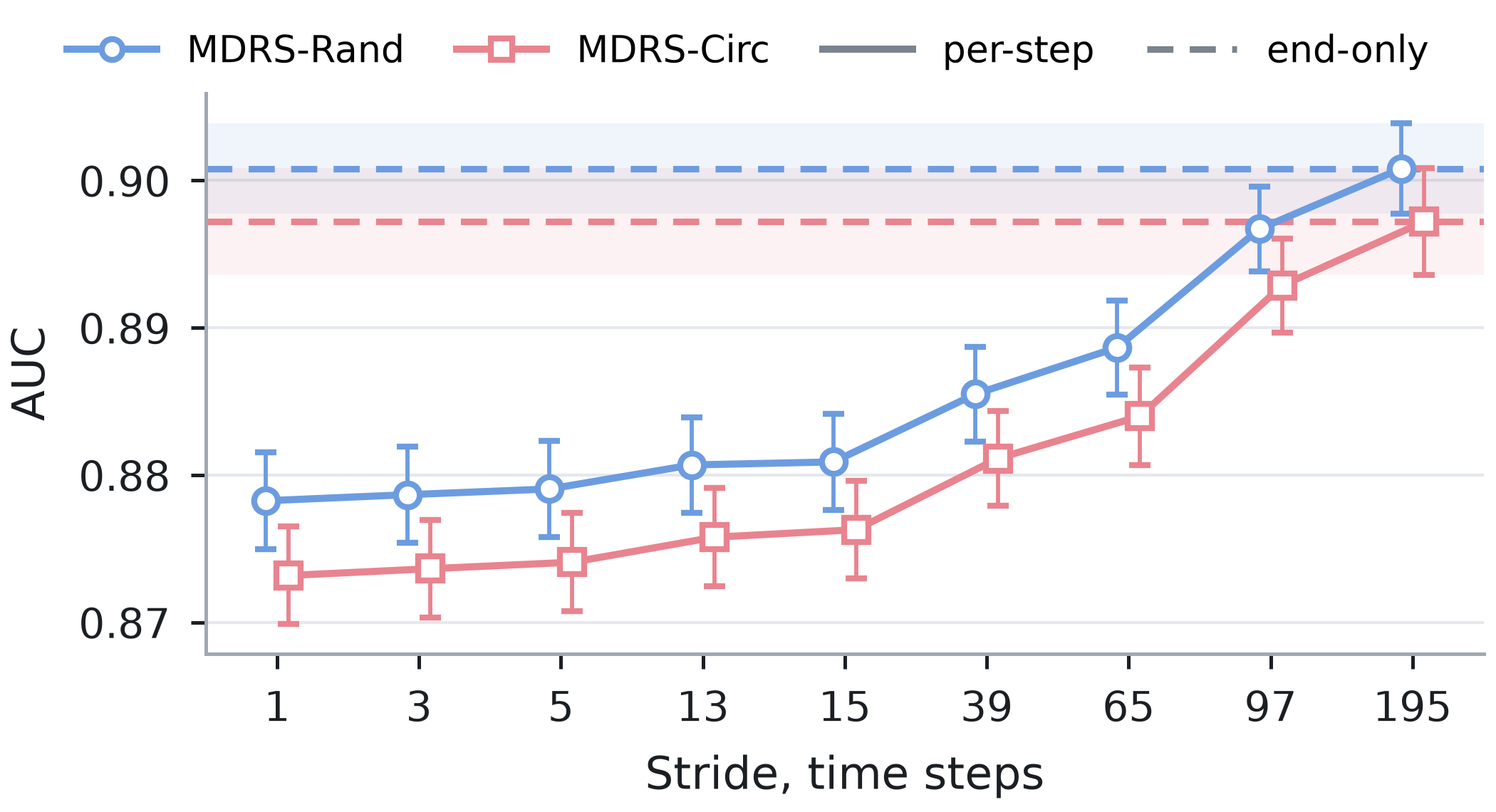}
    \caption{Test AUC of MDRS ($N=64$, MLPerf Tiny preprocessing) versus the stride of the per-step aggregation (solid lines) and with the end-only score (dashed lines), which coincides with stride 195. Mean over 50 reservoir initializations; bars and bands: 95\% confidence interval of the mean.}
    \label{fig:selected_points}
\end{figure}

Next, we investigate the effect of the aggregation strategies introduced in \Cref{sec:methods:aggregation}. 
\Cref{fig:selected_points} shows the average AUC on the test set over 50 reservoir initializations for the per-step aggregation with different strides (solid lines) and for the end-only score (dashed lines), for both \texttt{MDRS-Rand} and \texttt{MDRS-Circ} with $N = 64$.

The two connectivities behave almost identically: for every stride, their average AUCs differ by at most $0.005$. The AUC of the per-step aggregation increases with the stride, from $0.878 \pm 0.012$ and $0.873 \pm 0.012$ for a stride of 1 to $0.897 \pm 0.010$ and $0.893 \pm 0.011$ for a stride of 97 (\texttt{MDRS-Rand} and \texttt{MDRS-Circ}, respectively), and the end-only score, which corresponds to a stride of 195, achieves the highest AUC: $0.901 \pm 0.011$ and $0.897 \pm 0.013$.
A possible explanation is that per-step aggregation does not include a warm-up period.
The reservoir starts from a zero state and accumulates information about the recording gradually, so its early states are dominated by this transient and reflect only the beginning of the recording. The per-step aggregation weighs early states as much as later ones, whereas increasing the stride skips them, since the first scored state is at step $s$, and the end-only score uses only the final state, which contains information accumulated from the preceding frames. This is consistent with the AUC increasing with the stride and being highest for the end-only score.

Reducing the number of Mahalanobis distance evaluations per recording does not degrade the AUC; on the MLPerf Tiny benchmark, the final-state score achieves the highest anomaly detection performance.
Since this substantially reduces the computational cost of MDRS (\Cref{sec:exp:cost}), it is used in all subsequent experiments in this section.

\subsection{Comparison of computational cost}
\label{sec:exp:cost}
All proposed MDRS variants achieve the MLPerf Tiny target performance, allowing us to extend the analysis to their computational cost and memory footprint.

\subsubsection{Computational cost of MDRS variants}
We first investigate how the reservoir connectivity and the aggregation strategy affect the computational cost during inference. \Cref{tab:flops} reports the number of operations per recording for the four MDRS variants across different reservoir sizes.

\begin{table*}[!t]
\centering
\footnotesize
\setlength{\tabcolsep}{4pt}
\renewcommand{\arraystretch}{1.3}
\caption{Number of FLOPs (in millions) per recording during inference by the four MDRS variants as a function of reservoir size, $N$, measured with hardware performance counters. In parentheses: reduction with respect to \texttt{MDRS-Rand} with per-step aggregation.}
\label{tab:flops}
\begin{tabular}{l|rrrrrr}
  \textsc{Reservoir}, $N$ & $32$ & $64$ & $96$ & $128$ & $192$ & $256$ \\
  \midrule
  \texttt{MDRS-Rand}, per-step & 9.14 & 19.86 & 32.19 & 46.11 & 78.74 & 117.77 \\
  \rowcolor{gray!12}
  \texttt{MDRS-Rand}, end-only & 8.68 ($\downarrow$5.0\%) & 18.17 ($\downarrow$8.5\%) & 28.46 ($\downarrow$11.6\%) & 39.55 ($\downarrow$14.2\%) & 64.14 ($\downarrow$18.6\%) & 91.94 ($\downarrow$21.9\%) \\
  \texttt{MDRS-Circ}, per-step & 8.65 ($\downarrow$5.3\%) & 18.09 ($\downarrow$8.9\%) & 28.33 ($\downarrow$12.0\%) & 39.37 ($\downarrow$14.6\%) & 63.84 ($\downarrow$18.9\%) & 91.51 ($\downarrow$22.3\%) \\
  \rowcolor{gray!12}
  \texttt{MDRS-Circ}, end-only & 8.20 ($\downarrow$10.3\%) & 16.40 ($\downarrow$17.5\%) & 24.60 ($\downarrow$23.6\%) & 32.81 ($\downarrow$28.8\%) & 49.23 ($\downarrow$37.5\%) & 65.68 ($\downarrow$44.2\%) \\
  \bottomrule
\end{tabular}
\end{table*}

Replacing the random connectivity with the circular shift removes the $N\times N$ matrix-vector product $J\Phi(x(t))$ at every time step, whereas the input projection $Vf^{\mathrm{in}}(t)$, which dominates the cost of small reservoirs, is unchanged. The saving grows with reservoir size: with the per-step aggregation, \texttt{MDRS-Circ} requires 5.3\% fewer operations than \texttt{MDRS-Rand} for $N = 32$ and 22.3\% fewer for $N = 256$. Similarly, the end-only score avoids most of the Mahalanobis distance evaluations per recording, saving 5.0-21.9\% for \texttt{MDRS-Rand}. The two optimizations are complementary: \texttt{MDRS-Circ} with the end-only score requires 10.3-44.2\% fewer operations than \texttt{MDRS-Rand} with per-step aggregation. 

\subsubsection{Comparison to the baseline methods}
\begin{figure*}[t]
  \centering
  \includegraphics[width=\textwidth]{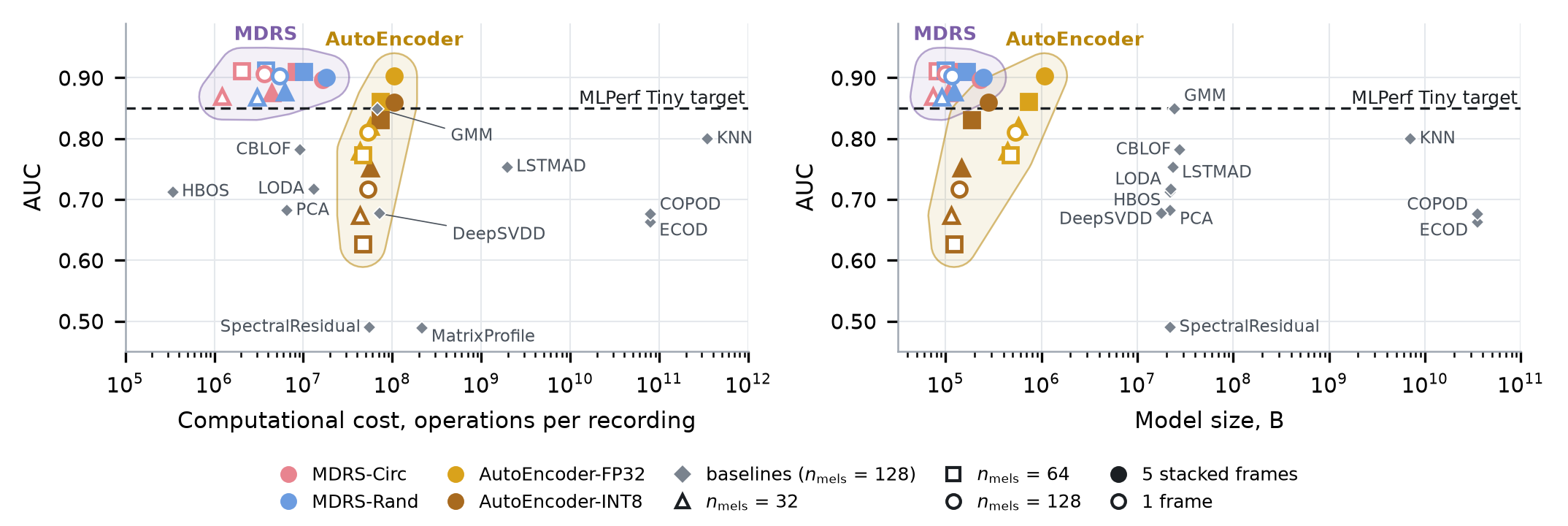}
  \caption{Anomaly detection performance versus computational cost (left) and model size (right) during inference. MDRS ($N=64$, averaged over $50$ reservoir initializations) and the autoencoder are shown for six preprocessing configurations: $n_\text{mels} \in \{32, 64, 128\}$ (marker shape), with five stacked frames (filled) or a single frame (hollow) per input sample. The baseline models use the MLPerf Tiny preprocessing.}
  \label{fig:model_performance_comparison}
\end{figure*}

\Cref{fig:model_performance_comparison} compares the proposed MDRS variants with the baseline methods (\Cref{sec:methods:baselines}) in terms of anomaly detection performance, computational cost during inference (left panel), and model size (right panel).  
We first consider the MLPerf Tiny preprocessing ($n_\text{mels}=128$, five stacked frames; \Cref{sec:methods:preprocessing}), shown as filled circles for MDRS and the autoencoder and used for all other baselines (gray diamonds). Four models achieve the target performance of $\text{AUC}\geq 0.850$: \texttt{MDRS-Rand} and \texttt{MDRS-Circ} ($0.901$ and $0.897$), \texttt{AutoEncoder-FP32} ($0.903$), and \texttt{AutoEncoder-INT8} ($0.860$), with MDRS requiring $18.2$ and $16.4$ million operations per recording, $5.8$ and $6.4$ times fewer than the autoencoder ($105.2$ million). None of the other baselines achieves the target among the evaluated configurations. The baselines requiring fewer operations than MDRS, \texttt{HBOS}, \texttt{PCA}, \texttt{CBLOF}, and \texttt{LODA} (0.34 to 13.0 million), achieve at most 0.781 (\texttt{CBLOF}). The best baseline overall, \texttt{GMM} with 128 diagonal Gaussian components, achieves 0.849 with $6.9\times10^{7}$ operations, about four times more than \texttt{MDRS-Circ}, and its parameters alone take about 650\,KB in single precision, more than the flash memory of the microcontroller used in Section~VI. \texttt{KNN} (0.800) requires $3.5\times10^{11}$ operations and 7.0\,GB, as it compares each test sample with all training samples. \texttt{LSTMAD}, which replaces the random reservoir with a trained recurrent network composed of LSTM cells, achieves 0.753 with $1.9\times10^{9}$ operations.

The MLPerf Tiny preprocessing, however, is tailored to the reference autoencoder: since the autoencoder has no memory and processes each input sample independently, the temporal context must be contained in the input sample itself, hence the stacking of five consecutive frames. The reservoir of MDRS, instead, integrates its input over time, so its state already retains information about the previous frames. We therefore evaluate MDRS also with a single frame per time step (hollow markers). Moreover, since MDRS exceeds the target performance by a wide margin with $n_\text{mels}=128$, we also evaluate fewer mel bands, $n_\text{mels}\in\{32, 64\}$ (triangles and squares), which further shorten the input sample. The autoencoder is evaluated with the same six preprocessing configurations, so that the two models are compared under the same preprocessing conditions. 
  
The single-frame input confirms the different role of the stacking in the two methods. For MDRS, it reduces the number of operations by a factor of $3.6-4.5$ while changing the AUC by less than $0.01$; for instance, with $n_\text{mels}=128$ it requires $3.6$ instead of $16.4$ million operations and achieves an AUC of $0.906$. For \texttt{AutoEncoder-FP32}, in contrast, the AUC decreases by $0.04-0.09$ and falls below the target performance (e.g., $0.810$ with $n_\text{mels}=128$), while the number of operations decreases by at most a factor of two, since the $128$-unit hidden layers dominate it. 
Reducing the number of mel bands lowers the cost of MDRS further while keeping it above the target. With $n_\text{mels}=64$ and a single frame, MDRS achieves its highest AUC ($0.912$ for both variants) with $2.0$ (\texttt{MDRS-Circ}) and $3.8$ (\texttt{MDRS-Rand}) million operations; with $n_\text{mels}=32$ and a single frame, \texttt{MDRS-Circ} requires only $1.2$ million operations, $13.4$ times fewer than with the MLPerf Tiny preprocessing, and still achieves an AUC of $0.869$; among all evaluated models, only \texttt{HBOS} requires fewer operations ($0.34$ million), at an AUC of $0.713$. The autoencoder, in contrast, loses AUC with fewer mel bands: with five stacked frames, \texttt{AutoEncoder-FP32} achieves $0.861$ with $n_\text{mels}=64$ and $0.820$ with $n_\text{mels}=32$, and \texttt{AutoEncoder-INT8} $0.831$ and $0.751$. Only \texttt{AutoEncoder-FP32} with $n_\text{mels}=64$ and five stacked frames achieves the target.

Finally, \Cref{tab:Peak_RAM_training_random_and_cyclic} reports the peak RAM of the models that achieve the target. Training MDRS requires 61\,MB and 2\,s (Intel Core i9-12900), five times less memory and about 1\,000 times less time than the autoencoder (315\,MB, 35\,min). During inference, MDRS requires about 700\,KB with the MLPerf Tiny preprocessing, between the INT8 (524\,KB) and FP32 (1\,168\,KB) autoencoders, and down to 372\,KB with $n_\text{mels}=32$ and a single frame. The memory of the baselines, none of which reaches the target, grows linearly with the training set; they require from 64\,MB to an estimated 40\,GB for training (\texttt{LSTMAD}: 39\,GB already with 700 recordings) and from 5.6\,MB to an estimated 65\,GB during inference.

Beyond its lower cost, training MDRS requires only normal data, which a device observes during regular operation without any labels or examples of faults, and involves no gradient computation: $\Sigma_0$ is a sum of rank-one terms that can be accumulated one reservoir state at a time without storing the training data, and $\Lambda_0$ can be updated directly with incremental methods, avoiding the matrix inversion. For $N=64$, $\Sigma_0$ occupies 16\,KB in single precision and fits in the RAM of the microcontroller used in \Cref{sec:exp:hard}. MDRS could therefore be trained on the device after installation and updated continuously during operation, e.g., to adapt to an individual machine or to slow changes in its normal behavior; we leave this to future work.

Overall, MDRS is the only model that achieves the target performance with every preprocessing, and with $n_\text{mels}=32$ and a single frame, \texttt{MDRS-Circ} requires the fewest operations, the smallest model, and the least memory during inference among the models that achieve the target.

\begin{table}[t]
\centering
\footnotesize
\setlength{\tabcolsep}{6pt}
\renewcommand{\arraystretch}{1.25}
\caption{Peak RAM during training (MB) and during inference (KB) for the models that achieve the target performance.}
\label{tab:Peak_RAM_training_random_and_cyclic}
\begin{tabular}{l|rr}
  \textsc{Model} & \textsc{Training (MB)} & \textsc{Inference (KB)} \\
  \midrule
  \texttt{MDRS-Rand}, $128\times5$ & 62 & 720 \\
  \rowcolor{gray!12}
  \texttt{MDRS-Circ}, $128\times5$ & 61 & 708 \\
  \texttt{MDRS-Rand}, $32\times1$ & 61 & 556 \\
  \rowcolor{gray!12}
  \texttt{MDRS-Circ}, $32\times1$ & 60 & 372 \\
  \texttt{AutoEncoder-FP32} & 315 & 1,168 \\
  \rowcolor{gray!12}
  \texttt{AutoEncoder-INT8} & 315 & 524 \\
  \bottomrule
\end{tabular}
\end{table}

\section{Experimental evaluation in hardware}
\label{sec:exp:hard}
To demonstrate the impact of MDRS variants under realistic resource-constrained deployment conditions, we ported the algorithm to the ST NUCLEO-F401RE board and compared it against the reference \texttt{AutoEncoder-INT8} in terms of three quantities that directly affect resource-constrained deployment: execution time, estimated energy consumption, and memory footprint. 
The NUCLEO-F401RE microcontroller features an $84$ MHz Cortex-M4F core with single-precision floating-point support, $512$ KB of flash memory, and $96$ KB of RAM. Running all kernels on identical hardware and using the same timing and energy-estimation procedure isolates the algorithmic differences while avoiding confounding board-dependent effects.

Based on the results of \Cref{sec:exp:soft}, we deploy MDRS with $n_\text{mels}=32$ and a single frame per time step. Among the evaluated preprocessing configurations, it requires the fewest operations and the smallest model, while still exceeding the target performance. To compare MDRS with the reference model under identical inputs, we also deploy it with the MLPerf Tiny preprocessing ($n_\text{mels}=128$, five stacked frames).

\subsection{Deployment and measurement setup}
\label{sec:hard:setup}

Following MLPerf Tiny, which excludes pre- and post-processing from the measurements, only the inference kernels run on the microcontroller; feature extraction, training, and the computation of the metrics are performed off-board. The board reads preprocessed feature vectors stored in flash, runs the timed kernel, and sends the anomaly score to the host, where it is compared with the Python reference. Since MLPerf Tiny defines one inference as one input window, we measure the execution time of one window with the Cortex-M4F cycle counter, discard five warm-up calls, and report the median of the following 21. With the per-step aggregation, a window includes one Mahalanobis distance evaluation; with the end-only score, only the reservoir update, since the final Mahalanobis distance, amortized over the 195 windows of a recording, adds less than 1\%. Energy per inference is estimated as the execution time multiplied by the run-mode power of the microcontroller from its datasheet (22.6\,mA at 3.3\,V), and the memory footprint as the flash footprint of the deployed model.

\begin{table}[t]
\centering
\footnotesize
\setlength{\tabcolsep}{3pt}
\renewcommand{\arraystretch}{1.25}
\caption{MDRS on the NUCLEO-F401RE ($N=64$, $32\times1$ input). Latency: median of 21 runs; parentheses: cycle reduction vs.\ per-step \texttt{MDRS-Rand}; AUC: mean of 50 initializations (std 0.001).}
\label{tab:hard:variants}
\resizebox{\columnwidth}{!}{%
\begin{tabular}{@{}l|rrrrr@{}}
  & \textsc{Cycles} & \textsc{Latency} & \textsc{Energy} & \textsc{Flash} & \\
  \textsc{Model variant} & (k) & (ms) & (\textmu J) & (KB) & \textsc{AUC} \\
  \midrule
  \texttt{MDRS-Rand}, per-step & 92.7 & 1.10 & 82 & 40 & 0.843 \\
  \rowcolor{gray!12}
  \texttt{MDRS-Rand}, end-only & 59.2 ($\downarrow$36\%) & 0.70 & 53 & 40 & 0.868 \\
  \texttt{MDRS-Circ}, per-step & 60.6 ($\downarrow$35\%) & 0.72 & 54 & 24 & 0.843 \\
  \rowcolor{gray!12}
  \textbf{\texttt{MDRS-Circ}, end-only} & \textbf{27.1 ($\downarrow$71\%)} & \textbf{0.32} & \textbf{24} & \textbf{24} & \textbf{0.869} \\
  \bottomrule
\end{tabular}}
\end{table}

\subsection{On-board characterization of MDRS variants}
\label{sec:hard:variants}

We first evaluate the four MDRS variants obtained by combining the two reservoir connectivities introduced in \Cref{sec:methods}: a random matrix $\mathbf{J}$ (\texttt{MDRS-Rand}) and the special case of permutation (circular shift, \texttt{MDRS-Circ}) with the two strategies of computing anomaly scores considered in \Cref{sec:methods:aggregation}: the per-step aggregation and the end-only score.

\Cref{tab:hard:variants} reports the on-board results for the four variants with the deployed preprocessing, $n_\text{mels}=32$ and a single frame. \texttt{MDRS-Rand} with the per-step aggregation requires $92.7$\,K-cycles per inference, corresponding to $1.10$\,ms and $82$\,\textmu J, and achieves an AUC of $0.843$, just below the target. Replacing the random connectivity with the circular shift removes the $N\times N$ matrix-vector product with $\mathbf{J}$, which reduces the cycles by 34.6\% and the flash footprint from 40 to 24\,KB. Replacing the per-step aggregation with the end-only score removes the $N\times N$ quadratic form of the Mahalanobis distance from every window, which reduces the cycles by 36.1\% and increases the AUC to $0.868$, above the target: with this preprocessing, the end-only score is necessary to meet the target. With 32-dimensional inputs, these two $N\times N$ operations dominate the cost of a window, so that their combination reduces the cycles by 70.7\%, consistent with the reduction in the number of operations reported in \Cref{sec:exp:cost} (74\%). Thus, \texttt{MDRS-Circ} with the end-only score requires only $27.1$\,K-cycles per inference, corresponding to $0.32$\,ms latency, and $24$\,\textmu J per inference, with a model of 24\,KB.

\subsection{Iso-hardware comparison with the reference model}
\label{sec:hard:ae}

We next compare the most optimized MDRS implementation against the fully-connected autoencoder adopted as the MLPerf Tiny reference model. A direct comparison with publicly available MLPerf Tiny submissions is affected by hardware variability: different submissions use different microcontrollers, clock frequencies, memory systems, DSP extensions, and power domains. This makes it difficult to separate algorithmic efficiency from board-dependent effects. To remove this confounding factor, we ported the baseline \texttt{AutoEncoder-INT8} to the same NUCLEO-F401RE microcontroller used for investigating MDRS variants and measured it with the procedure described above.

\begin{table}[t]
\centering
\footnotesize
\setlength{\tabcolsep}{3pt}
\renewcommand{\arraystretch}{1.3}
\caption{Autoencoder and end-only \texttt{MDRS-Circ} on the NUCLEO-F401RE.}
\label{tab:hard}
\resizebox{\columnwidth}{!}{%
\begin{tabular}{l|rrrrr}
  \textsc{Model} & \textsc{Cyc.~(k)} & \textsc{Latency} & \textsc{Energy} & \textsc{AUC} & \textsc{Flash} \\
  \midrule
  \texttt{AutoEncoder-INT8}, $128\times5$ & 2\,058.7 & 24.51\,ms & 1\,828\,\textmu J & 0.860 & 270\,KB \\
  \rowcolor{gray!12}
  \texttt{MDRS-Circ}, $128\times5$ & 297.9 & 3.55\,ms & 264\,\textmu J & 0.897 & 176\,KB \\
  \textbf{\texttt{MDRS-Circ}, $32\times1$} & \textbf{27.1} & \textbf{0.32\,ms} & \textbf{24\,\textmu J} & \textbf{0.869} & \textbf{24\,KB} \\
  \midrule
  \textsc{Ratio} (\texttt{AE}\,$\div$\,MDRS), $128\times5$ & $6.9\times$ & $6.9\times$ & $6.9\times$ & --- & $1.5\times$ \\
  \rowcolor{gray!12}
  \textsc{Ratio} (\texttt{AE}\,$\div$\,MDRS), $32\times1$ & $75.9\times$ & $75.9\times$ & $75.9\times$ & --- & $11.3\times$ \\
  \bottomrule
\end{tabular}%
}
\end{table}

\Cref{tab:hard} reports the results. With the same input samples, i.e., the MLPerf Tiny preprocessing, \texttt{MDRS-Circ} requires $297.9$\,K-cycles per inference, corresponding to $3.55$\,ms and $264$\,\textmu J, while achieving an AUC of $0.897$. On the same microcontroller, \texttt{AutoEncoder-INT8} requires $2\,058.7$\,K-cycles, $24.51$\,ms, and $1\,828$\,\textmu J to reach an AUC of $0.860$. Under iso-hardware settings and identical inputs, MDRS is thus $6.9\times$ faster and more energy-efficient than the MLPerf Tiny reference model, while improving the AUC and reducing the memory footprint by $1.5\times$.

With the deployed preprocessing, the advantage grows to $75.9\times$ in time and $11.3\times$ in memory footprint, while MDRS still exceeds the target performance (AUC $0.869$), whereas \texttt{AutoEncoder-INT8} reaches it only with the MLPerf Tiny preprocessing (\Cref{sec:exp:cost}). This gain combines two effects: the lower inference cost of MDRS for the same input samples ($6.9\times$), and the ability of MDRS to operate on much shorter input samples, which reduces its own cost by a further $11.0\times$. 
With a new input window every 32\,ms (the front-end hop), the autoencoder uses 77\% of the real-time budget, whereas MDRS uses 11\% with the MLPerf Tiny preprocessing and 1\% with the deployed one, leaving the microcontroller almost entirely free for the front-end and other tasks.

\section{Conclusion} 
Anomaly detection on resource-constrained IoT devices requires meeting detection performance targets while minimizing computational cost and memory footprint. In this study, we investigated the suitability of MDRS method for anomaly detection on resource-constrained IoT devices using the MLPerf Tiny benchmark. We proposed two algorithmic optimizations to improve its efficiency: circular shift for reservoir connectivity (\texttt{MDRS-Circ}) and end-only anomaly scoring. Together, they reduce inference operations by up to $44.2\%$ compared with standard MDRS, while maintaining or improving detection performance.

When compared with several other anomaly detection methods, MDRS and the reference autoencoder are the only methods meeting the MLPerf Tiny performance target of $\text{AUC}\geq 0.850$.  
With MLPerf Tiny preprocessing, the optimized MDRS variant achieves an AUC of $0.897$ with $6.4\times$ fewer operations than the autoencoder. Exploiting the reservoir's memory capacity enables single-frame inputs with $n_\text{mels}=32$, achieving an AUC of $0.869$ with only $1.2$ million operations per recording ($13.4\times$ fewer than with the standard preprocessing), whereas the autoencoder falls below the target with such preprocessing. 

Deployment on the NUCLEO-F401RE microcontroller confirms these improvements. MDRS achieves $0.32~\mathrm{ms}$ latency, $24~\mu\mathrm{J}$ per inference, and a $24~\mathrm{KB}$ model, corresponding to $75.9\times$ lower latency and estimated energy consumption and an $11.3\times$ smaller model than the autoencoder using the standard preprocessing. 
These results demonstrate the potential of reservoir computing for efficient anomaly detection on resource-constrained IoT devices. Although the benchmark uses acoustic recordings, MDRS processes their log-mel features as a multivariate time series without audio-specific assumptions, making other sensor modalities a natural direction for future evaluation. Future work will also extend MDRS to other tasks, evaluate on-device training and continual adaptation through incremental updates of the precision matrix, and investigate further algorithmic and hardware-specific optimizations.

\bibliographystyle{IEEEtran} 
\balance
\bibliography{References}

\end{document}